\documentclass[preprint]{elsarticle}
\usepackage{graphicx, amssymb, subfigure}
\journal{Applied Mathematics and Computation}
\begin{document}
\begin{frontmatter}

\title{A Variation of the Box-Counting Algorithm Applied to Colour Images}

\author[TEI,gen]{N.~S.~Nikolaidis\corref{cor}}
\cortext[cor]{Corresponding author.}
\ead{niknik@teithe.gr}

\author[ece]{I.~N.~Nikolaidis}
\ead{iwavvns@gmail.com}

\author[gen]{C.~C. Tsouros}
\ead{tsouros@gen.auth.gr}

\address[TEI]{Department of Automation, Faculty of Applied Technology, Alexander Technological Educational Institute (ATEI) of Thessaloniki, GR-57400 Thessaloniki, Greece.}

\address[gen]{School of Mathematics, Physics and Computational Sciences, Faculty of Engineering, Aristotle University of Thessaloniki, GR-54124 Thessaloniki, Greece.}

\address[ece]{Department of Electrical and Computing Engineering, Aristotle University of Thessaloniki, GR-54124, Thessaloniki, Greece.}

\begin{abstract}
The box counting method for fractal dimension estimation had not been applied to large or colour images thus far due to the processing time required. In this letter we present a fast, easy to implement and very easily expandable to any number of dimensions variation, the box merging method. It is applied here in RGB images which are considered as sets in 5-D space.
\end{abstract}

\begin{keyword}
box counting, fractal dimension, fractals, image classification, image processing
\PACS 07.05.Pj, 07.05.Kf
\end{keyword}

\end{frontmatter}

\section{Introduction}\label{sec:Intro}
Hausdorff in 1918 \cite{Hausdorff1918}, in cooperation with Besicovitch, calculated, for the first time, the dimension of some irregular sets as a fractional number. Richardson, in his attempt to measure the real length of the borders between countries, presented in 1961 \cite{Richardson1961} the relation $L(G)=M G^{1-D}$, where $L$ is the actual length of a line, $G$ is the measurement scale, $M$ is the measure and $D$ is a constant. Mandelbrot, in 1967, realized that geographic borders can have a property of statistical self-similarity and that the exponent $D$ measures the Hausdorff dimension of the border. The meaning of fractals and the physical meaning of $D$ though, was presented systematically for the first time by Mandelbrot in 1977 in his famous book "The fractal geometry of nature" \cite{Mandelbrot1977}.

Mandelbrot defined a set as a fractal if it is a union of $n$ different subsets, each one a copy of the original in a scale $r<1$ in all coordinates. He defined the fractal dimension $D$ by the relation

\begin{equation} \label{eq:Ddefinition}
1=n \cdot r^D .
\end{equation}

In order to measure $D$, one can use equation

\begin{equation} \label{eq:Dmeas}
D=\frac{\log n}{\log s}
\end{equation}

where $s=1/r$. In order to apply (\ref{eq:Dmeas}) to fractal dimension measuring, one can use the box counting (BC) algorithm, introduced by Gagnepain and Roques-Carmes in 1986 \cite{Gagnepain1986} as "reticular cell counting". It works by iteratively scanning the volume $L_x \times L_y \times L_z$ in which the set is contained, with a box of size $\varepsilon_x \times \varepsilon_y \times \varepsilon_z = \frac{L_x}{s} \times \frac{L_y}{s} \times \frac{L_z}{s}$ and counting of the number $n$ of non-empty boxes. The scan is carried out with non-overlapping boxes for $s=2^\nu$, where $\nu$ is the iteration number, computing $n$ for each $s$. The procedure stops when $\varepsilon$ is less than or equals the size of the lattice on which the set is built. The fractal dimension can be calculated from the slope of the $\log n - \log s$ plot.

The method of mass distribution for fractal measurement was presented by Voss in 1986 \cite{Voss1986} and implemented by Keller et al in 1989 \cite{Keller1989} for image segmentation. The box-size values are, according to this method, not standard but all the possible ones from 1 to $L$. Also, the box does not have predetermined grid positions but it centers in every data element of the 3-D space. For every box size and every box position, the number of data elements $m$ inside the box is counted (the mass). The number of boxes $n$ for each $\varepsilon$ is computed from the counting of all the boxes of size $\varepsilon$ with $m=1$ pixel inside, with $m=2$, $m=3$, ... $m=L^3$. This method is more accurate and can easily compute local fractal dimension but it requires much more processing power.

The binary concatenation variation of box counting was proposed by Liebovitch and Toth in 1989 \cite{Liebovitch1989} and is as follows: For each element of the set, a string is created by concatenating the binary representations of its normalized coordinates on all the axes of the space. For each iteration, a logic "and" is performed with a string formed by 1's for the first $\nu_{max}-\nu$ digits and 0's for the rest, where $\nu_{max}$ is the total iteration count. After that, these numbers are sorted and the number of changes between them is counted. The fractal dimension is computed as usual.

The gliding box method introduced by Allain and Cloitre in 1991 \cite{Allain1991} uses a box of size $\varepsilon$ which is successively centered in all pixel positions. The mass $m$ is measured and the probability distribution $Q(m,\varepsilon)$ is obtained such that a gliding box of size $\varepsilon$ contains mass $m$. After computing the moments of this distribution, both $D$ and the lacunarity can be computed.

The method of differential box counting (DBC) presented by Sarkar and Chaudhuri in 1992 \cite{Sarkar1992} substitutes the time-consuming scanning in vertical axis with a simple subtraction. Instead of scanning the whole 3-D space in which the set is contained with a cube, one scans the x-y plane with a square and measures the height between the maximum and minimum values in z axis; this is assumed to be the total number of boxes found over the square. This method only gives correct results if the set is dense and uniformly distributed in vertical axis. However, if, at the same column, only the topmost and lowermost partitions are non-empty, the DBC algorithm will measure $L_z$ in box units instead of 2. Another drawback is that it is not easily expandable to more than 3 dimensions; it is theoretically possible, but implementing it was not deemed worth the effort.

The successive partitioning algorithm, presented by Molteno in 1993 \cite{Molteno1993}, involves dividing each box into $2^{\nu E}$ sub-boxes, disregarding all empty boxes for the rest of the process, thus saving processing time.

A method of measuring of fractal dimension of colour images was made by Lindstrom, in 2008 \cite{Lindstrom2008}. The method used is an extension of the variation method (essentially a "sliding differential box counting") in many dimensions. However, the proposed method needs more than 3 colour channels to be accurate and the computational resources it requires are very high.

An extension to the above-mentioned mass distribution method in colour space has been done recently in an unpublished work of Ivanovici and Richard \cite{Ivanovici2010}. For a certain square of size $\varepsilon$ in the $x-y$ plane, they count the number of data elements (pixels) that fall inside a 3-D cube of size $\varepsilon$ centered in the current pixel.

\section{The proposed method}\label{sec:ThePropMeth}
For a variety of reasons, no "proper" box counting algorithms had been systematically applied to images thus far. Instead, approximations such as DBC were commonly employed. In the present paper, it was decided that no such approximations were necessary: the method used is fast, reasonably accurate and easily expands to any number of dimensions. In particular, any new dimension the set might obtain merely adds a new column in a "partition table", whose creation is detailed below. From it, $n$ can be easily computed, and then $D$ as usual.

In our method, no box is being moved around per se. Instead, each axis of the image is first partitioned into $s$ partitions to create an E-dimensional grid. We start with the finest possible partitioning,

\begin{equation} \label{eq:smax}
s_{max}=L_{min}=min(L_x, L_y, L_z)~.
\end{equation}

The iterations will take place for the values of $s=s_{max}/2^\nu, \nu=\nu_{max},\nu_{max}-1,...,1$, where

\begin{equation} \label{eq:numax}
\nu_{max}= \lfloor log_2(L_{min}) \rfloor
\end{equation}

is the maximum number of iterations if we halve the edges $\varepsilon$ of the box in each iteration according to relation

\begin{equation} \label{eq:eLs}
\varepsilon = \frac{L}{s}
\end{equation}

for each axis.

The partitioning on axis x will then be determined by

\begin{equation} \label{eq:part}
t_x = \lfloor \frac{x}{\varepsilon_x} \rfloor = \lfloor \frac{x ~ s}{L_x} \rfloor
\end{equation}

\noindent where $x$ is the coordinate of any pixel in the box and $t_x$ is the coordinate of the box. The same normalization applies for all axes. This means that, if we know the $(x,y)$ coordinates of the pixel we can find the box that this pixel belongs in, keeping in mind that $z=f(x,y)$.

In box merging, the first iteration is executed with the finest possible partitioning ($s=s_{max}$). The key idea is that, if the positions of its non-empty boxes are known, the non-empty boxes belonging to the immediately coarser partitioning ($s_{max}/2$) can be found without further scanning. So, we construct a table $\it{A}$ with the coordinates of all the partitions which contain at least one element of the data set. The partition table of a new iteration can easily be produced by an integer division by 2 of all the contents of $\it{A}$ and merging of the identical rows, so no scanning is needed any more.

The algorithm\footnote{A Matlab version of the box merge algorithm can be downloaded from address http://www.autom.teithe.gr/niknik/FDBoxMerge.m}
 is as follows.

\noindent $s \leftarrow L_{min}$ (\ref{eq:smax})\\
Calculate $\varepsilon$ for all axes (\ref{eq:eLs})\\
For all pixel coordinates x,y\\
\indent Compute partition coordinates $t$ (\ref{eq:part})\\
\indent Put the $t$ values in a new row in partition table $\it{A}$\\
End for\\
Eliminate identical rows of $\it{A}$\\
Store the number $n$ of rows of $\it{A}$ and the current value of $s$.\\
Calculate maximum number of iterations (\ref{eq:numax})\\
For all iterations but the previous one\\
\indent Divide the contents of $\it{A}$ by 2 and take the integer part\\
\indent Eliminate identical rows to create the new partition table $\it{A}$\\
\indent Store the number $n$ of rows of $\it{A}$ \\
\indent Divide $s$ by 2 and store the new value\\
End for\\
Calculate $D$ from stored values of $n$ and $s$ (\ref{eq:Dmeas}) and (\ref{eq:lognmax})\\

\section{Calculation of fractal dimension from $\log~n - \log~s$ plot} \label{sec:Evaluation}
\begin{figure}[ht]
  \centering
  
  \subfigure[Log n -log s plot with resolution as a parameter.]{
  \includegraphics[trim=6mm 2mm 13mm 8mm, clip, width=4.1 cm, height=4.1 cm] {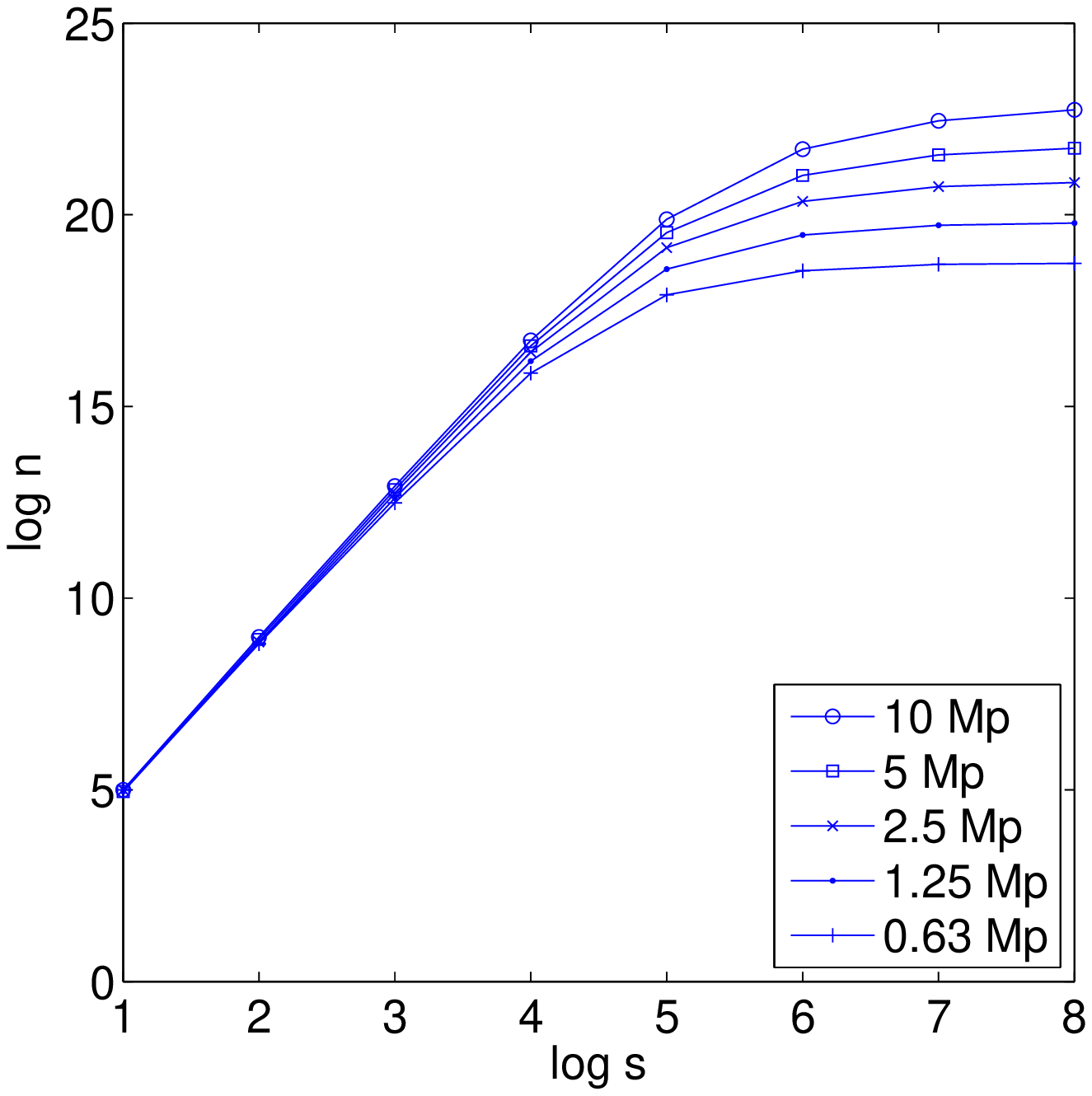}
  \label{fig:ResolTest}
  }
  \subfigure[Log n -log s plot with blur radius as a parameter]{
  \includegraphics[trim=6mm 2mm 13mm 8mm, clip, width=4.1 cm, height=4.1 cm]{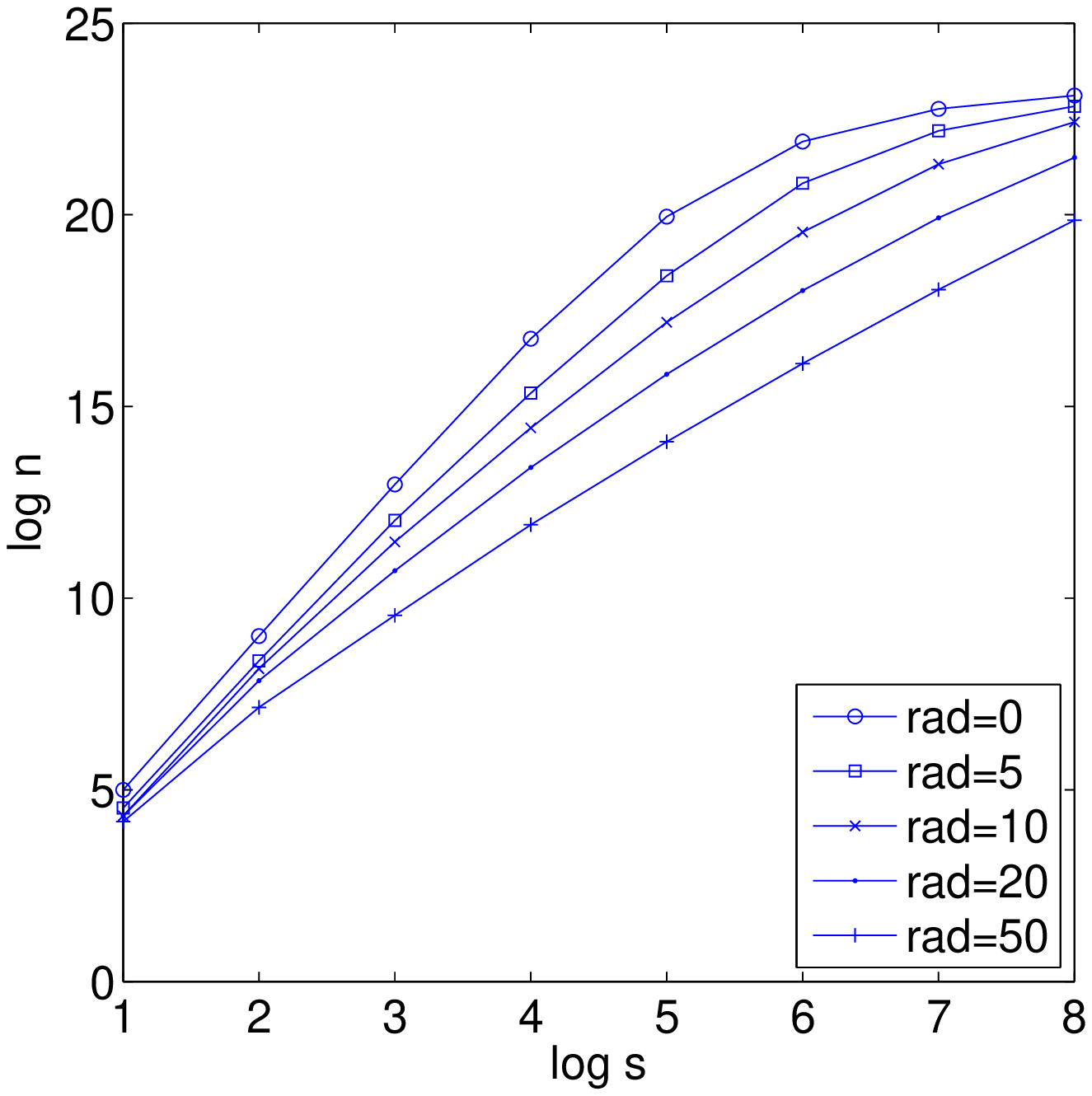}
  \label{fig:BlurTest}
  }
  \caption{Resolution and Blur tests.}
  \label{fig:ResolBlurTests}

\end{figure}

In order to better understand the way the shape of the $\log n - \log s$ plot affects the calculation of $D$ we selected a picture with high self-similarity (a scene in a forest) and calculated the log - log plots in various resolutions (Fig \ref{fig:ResolTest}) and various blur radii (Fig \ref{fig:BlurTest}). It can be seen that, when the resolution falls, $D$ remains almost constant whereas it drops when the image is being blurred. It can easily be observed that, when the fractal dimension is high and the size of the picture is limited, the plot abruptly turns to a horizontal line at about the height that corresponds to the value $\log_2(L_x ~ L_y)$. This is obviously caused by the fact that the picture is only finitely divisible; the boxes that the algorithm counts could never be greater than the total amount of pixels in the picture. Thus, even though the plot rises straightly at first, there is a cut-off at a predetermined height. Taking this cut-off into account in calculating $D$ will produce wrong results.
If the horizontal part of the plot corresponds to the logarithm of the size, it means that this image needs more pixels to express its self-similarity. In other words, the image has too high a fractal dimension to be kept in this size. To counteract that, it was decided that all the values of $log~n$ approximately equal to the logarithm of the size of the image must be disregarded, i.e.

\begin{equation} \label{eq:lognmax}
\left(\log_2 n\right)_{max} = 0.9~\log_2(L_x~L_y)~.
\end{equation}

The cut-off of 0.9 is a heuristic one and proved to give satisfactory results in all situations. Box dimension is calculated from the slope of the linear fit of the non-rejected values. 


\section{Evaluation of the algorithm} \label{sec:Check}
\begin{figure}[ht]
  \centering

  \subfigure[ A line (x, y, r, g, b) = (0, 0, 255, 0, 0) - (255, 255, 0, 255, 255) (D=1)] {
  \includegraphics[trim=17mm 21mm 26mm 16mm, clip, width=4cm, height=4cm]{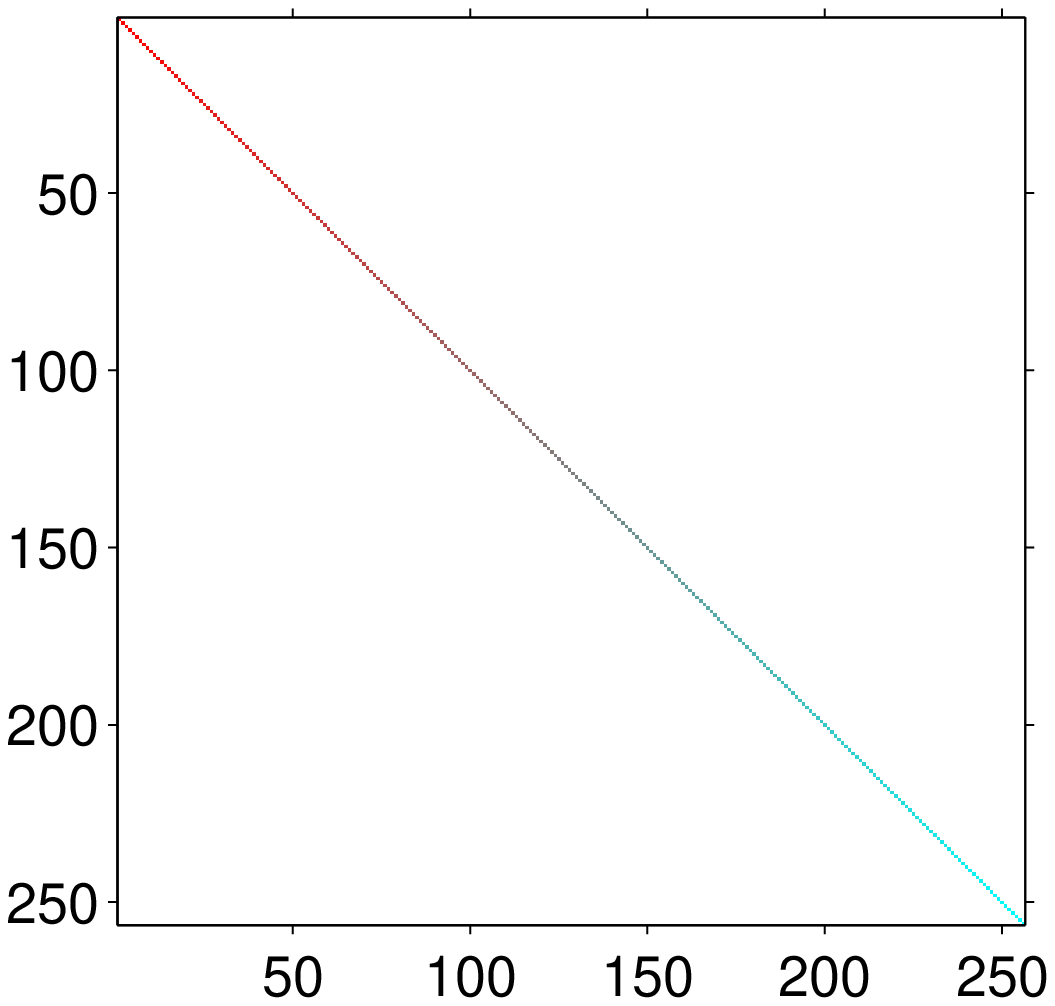}
  \label{fig:ColoredLine}
  }
  \subfigure[Gradient in R and G, B=128 (D=2)] {
  \includegraphics[width=4cm, height=4cm]{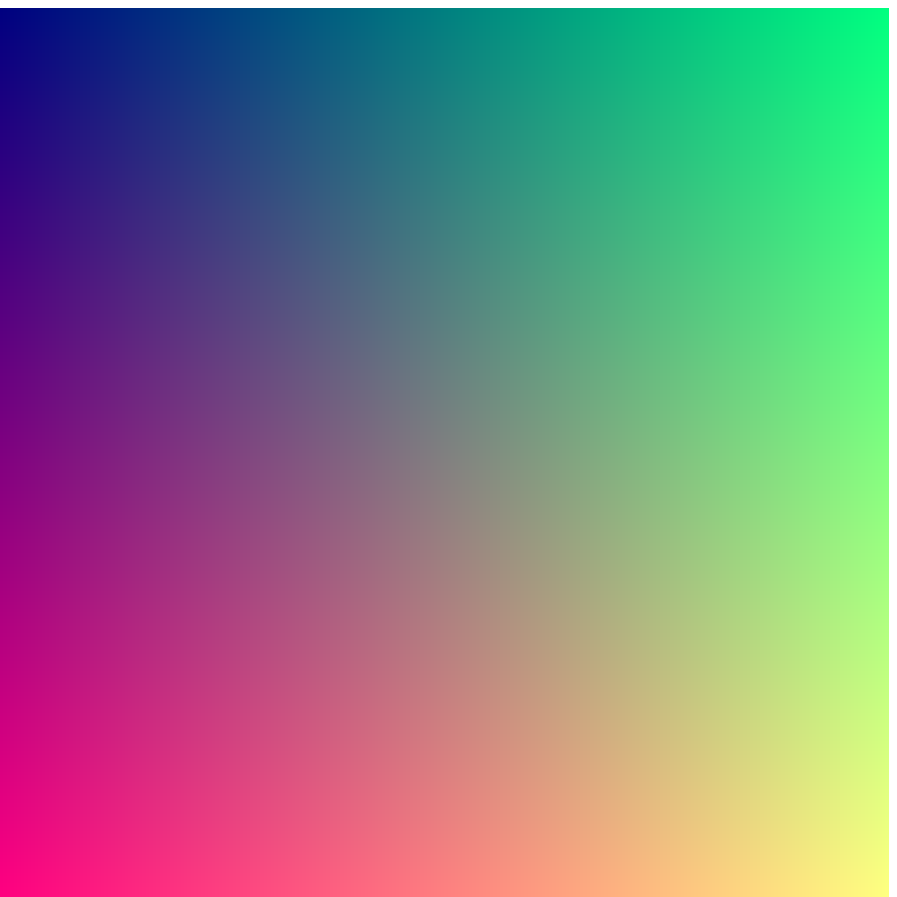}
  \label{fig:noise0}
  }
  \subfigure[Gradient in R and G, noise on B (D=3)] {
  \includegraphics[width=4cm, height=4 cm]{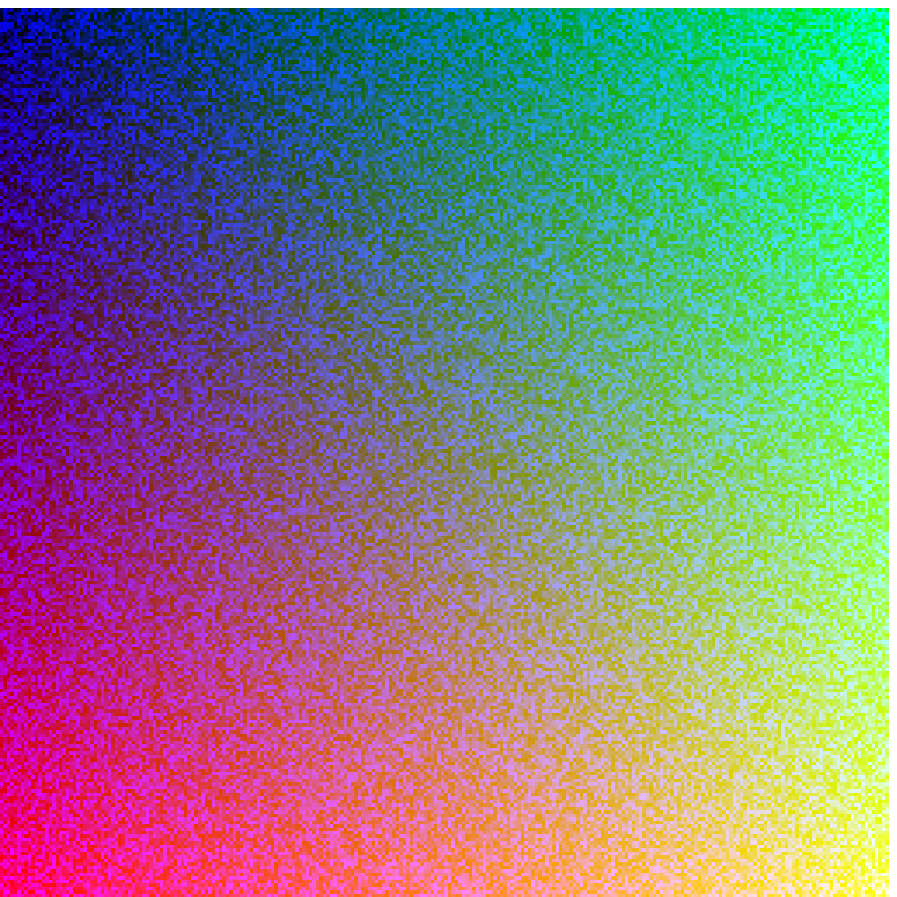}
  \label{fig:noise1}
  }
  \subfigure[Gradient in R, noise on G and B (D=4)] {
  \includegraphics[width=4cm, height=4cm]{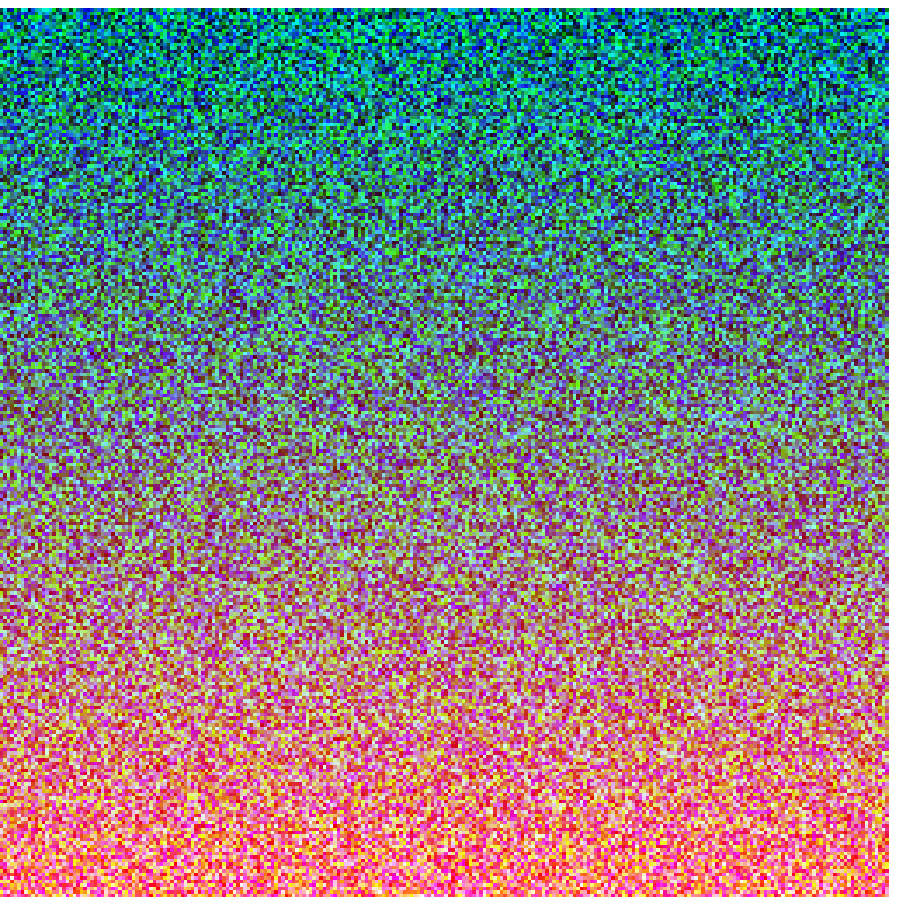}
  \label{fig:noise2}
  }
  \subfigure[Colour noise (D=5)] {
  \includegraphics[width=4cm, height=4 cm]{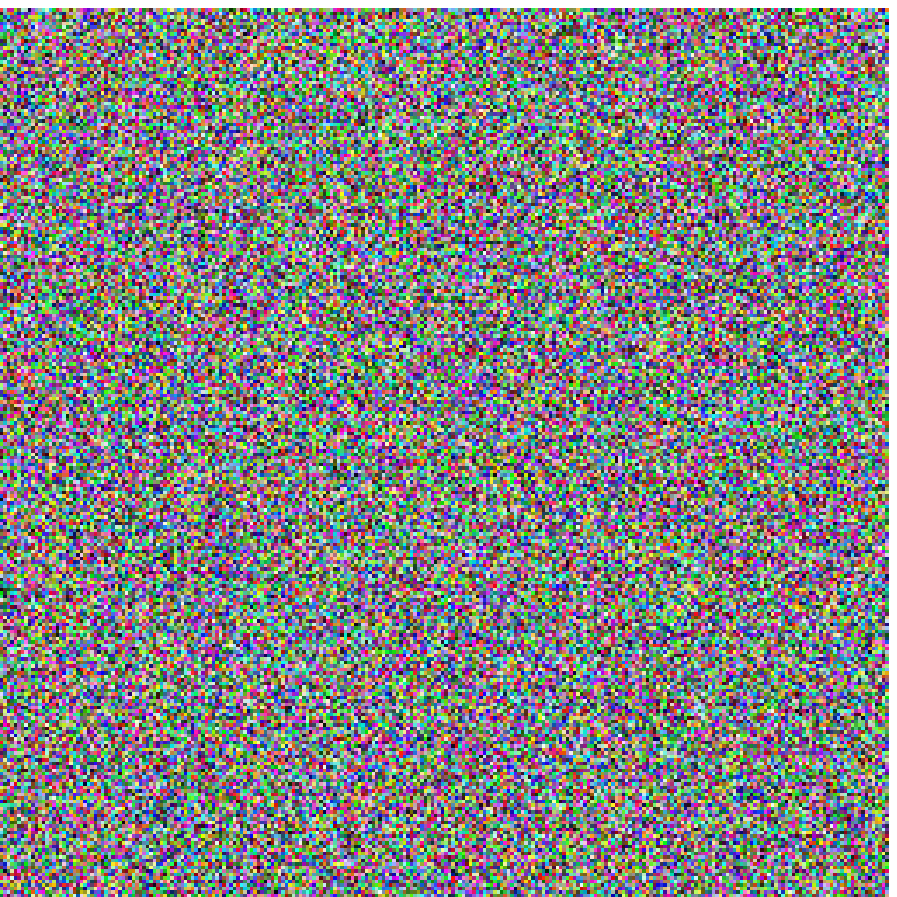}
  \label{fig:noise3}
  }
  \subfigure[ $\log n - \log s$ plots of the test images] {
  \includegraphics[trim=6mm 2mm 11mm 9mm, clip, width=4cm, height=4cm]{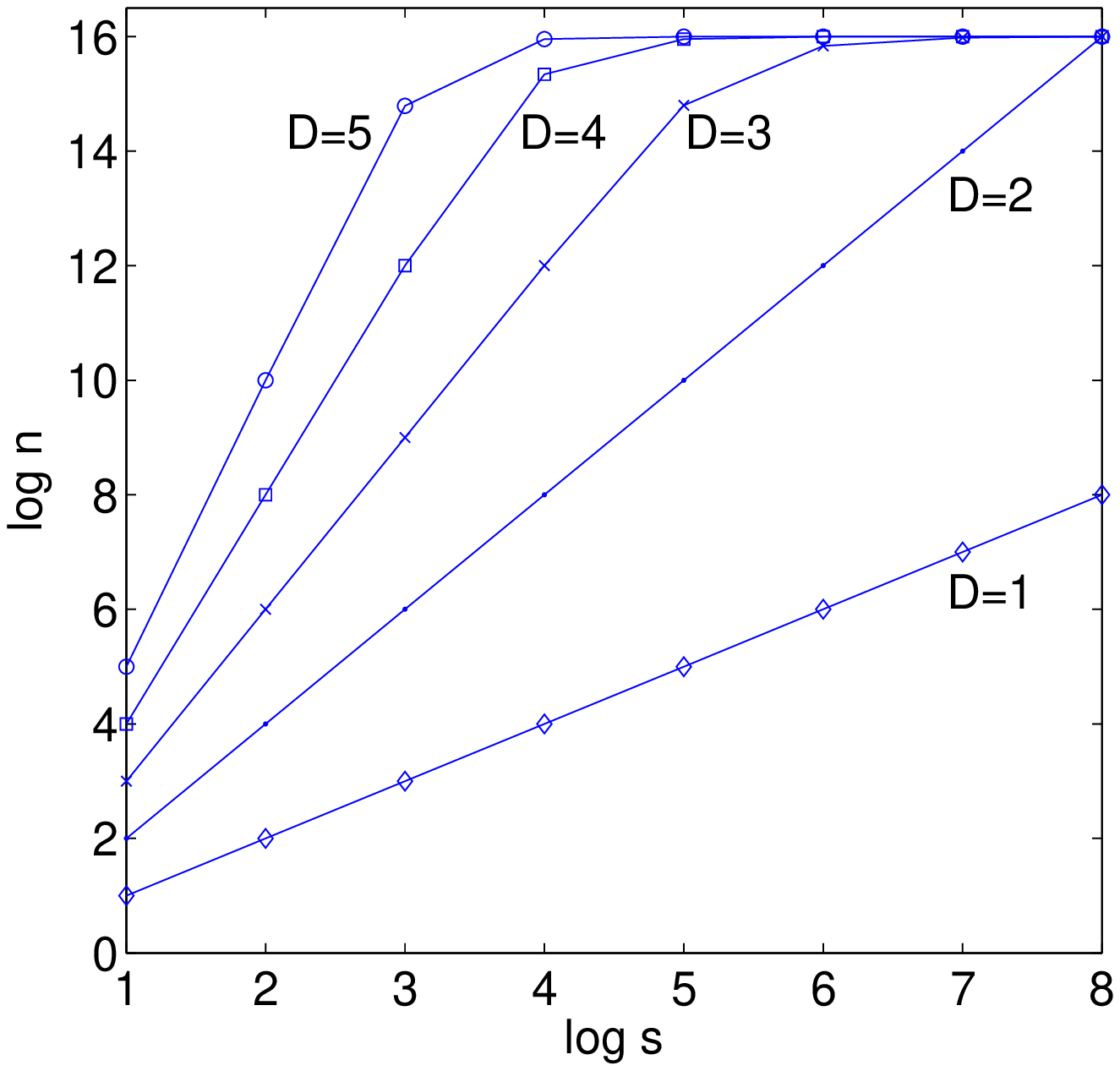}
  \label{fig:PlotsAll}
  }
  \caption{Test images.}
  \label{fig:TestImages}
\end{figure}

The colour image is a mapping from a 2-D space of pixel positions to the 3-D space of RGB colours. Because the entire image frame in the $x-y$ plane is used, the minimum fractal dimension will be $D_{min}=2$. On the other end, fractal dimension will always be smaller than the Euclidean dimension of the space where the data set is contained, which means that $D_{max}=5$. To create a shape with a fractal dimension smaller than 2, there would need to be an alpha channel that would define some pixels as transparent, disregarding their colour information. In such a situation, the fractal dimension of the colour image can become as low as 0. For example, we can define a coloured line with fractal dimension 1, where all of the pixels not belonging to it are transparent;  this means that the partition table will have less than $L_x \times Ly$ rows (pixels). As a first test, we have created a line in the 5-D image space from (x,y,r,g,b)=(0,0,255,0,0) to (255,255,0,255,255) (Fig \ref{fig:ColoredLine}) and checked its fractal dimension; the program gave us D=1 exactly (Fig \ref{fig:PlotsAll}).

As a second test, we have created a set of four images with size $256 \times 256$ pixels and known fractal dimensions. The first image (Fig. \ref{fig:noise0}) has been formed by a plane in 5-D space. Red channel has a gradient 0-255 along the x axis, green a gradient 255-0 along the y axis, and blue channel has constant value 128. As expected, the fractal dimension was measured $D=2$ exactly (Fig. \ref{fig:PlotsAll}). The second image is a copy of the previous one, except that random numbers were placed in blue channel instead. Its measured fractal dimension was $D=3$. The third image has random values in green as well, and the last one has random numbers in all three channels; the measured values were $D=4$ and $D=5$ respectively, as expected. Of course, colour noise in all channels produces the earliest cut-off at all resolutions.

\begin{figure}[ht]
  \centering
  \subfigure[] {
  \includegraphics[width=4cm]{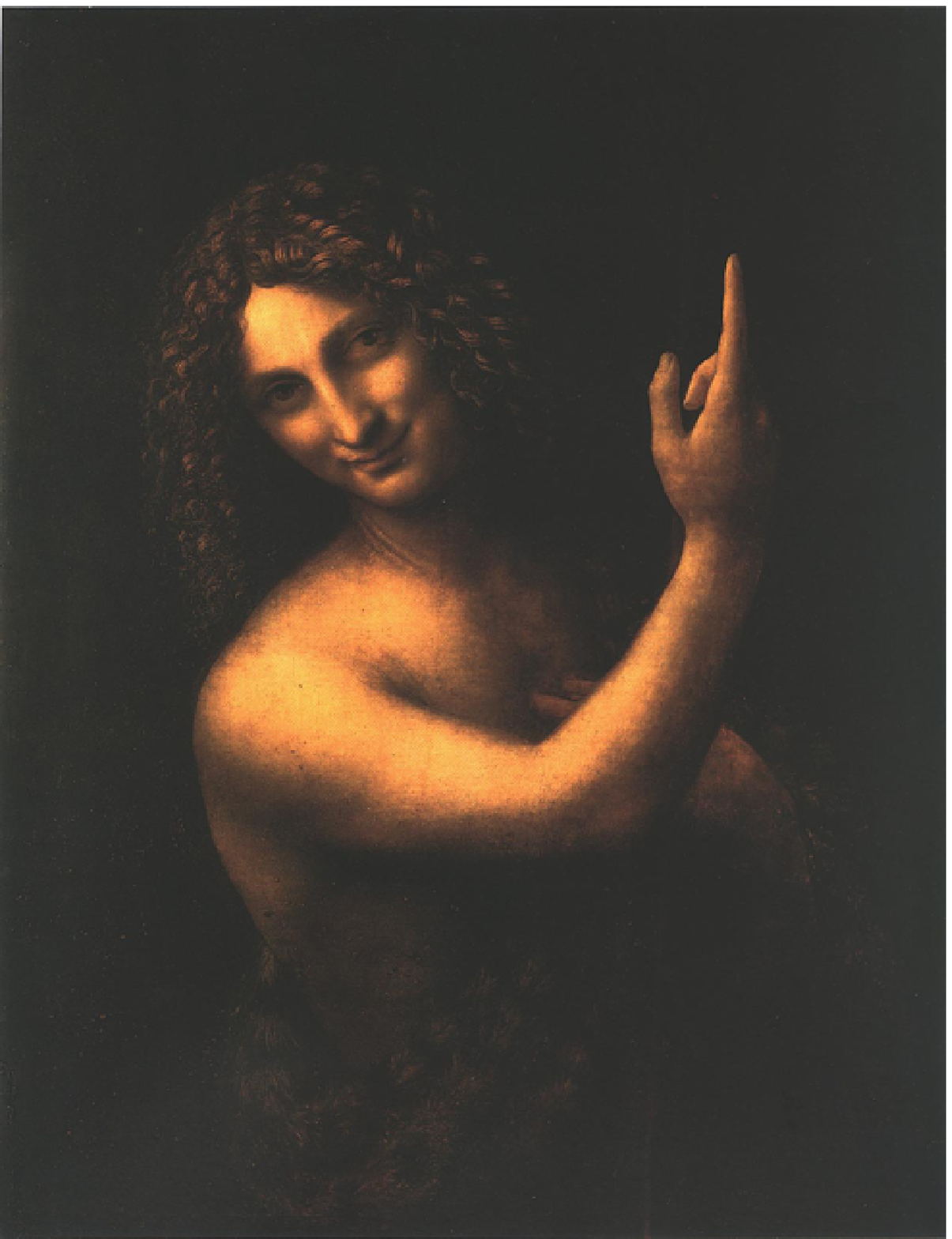}
  \label{fig:LdV}
  }
  \subfigure[]{
  \includegraphics[width=4cm]{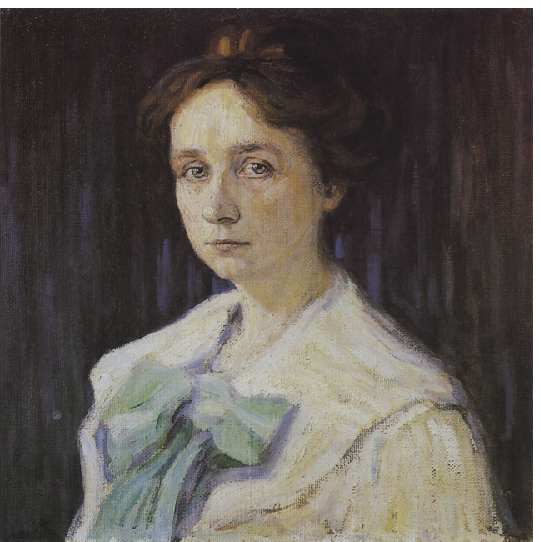}
  \label{fig:Kandinski}
  }
  \subfigure[] {
  \includegraphics[width=4cm]{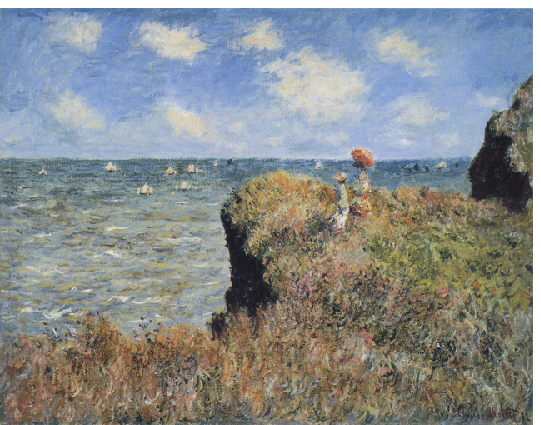}
  \label{fig:Monet}
  }
  \subfigure[] {
  \includegraphics[width=4cm]{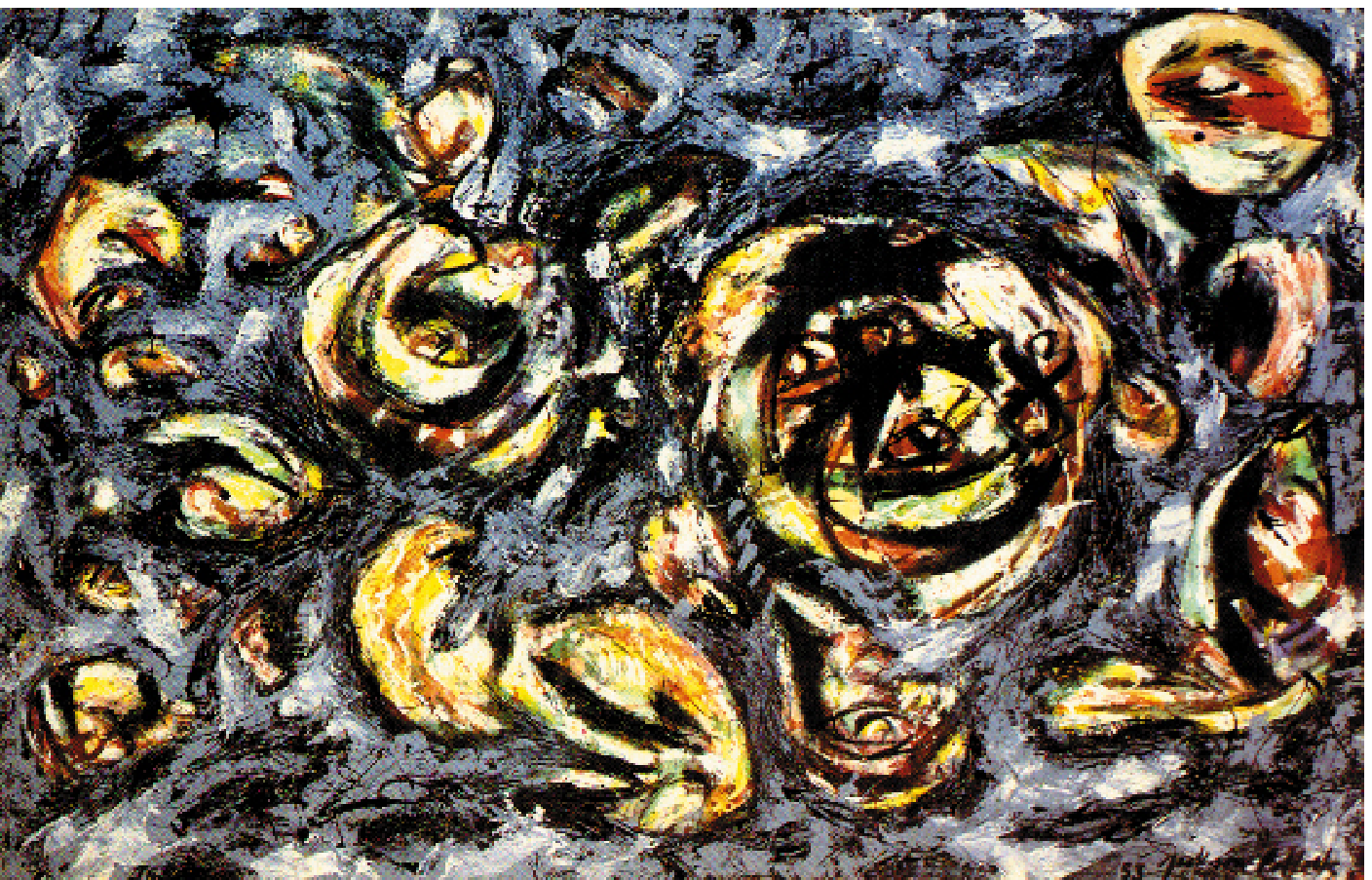}
  \label{fig:Pollock}
  }
  \caption{Measures of a sample of four paintings. (a) Leonardo da Vinci D=2.45 (b) Kandinski D=3.25 (c) Monet D=3.66 (d) Pollock D=4.18}
  \label{fig:PaintingsTest}
\end{figure}

As an application of the box merging algorithm, the fractal dimension of numerous known paintings was measured. A representative sample of the results, with low, medium and high $D$ can be seen in Fig. \ref{fig:PaintingsTest}. The biggest box dimension was measured (not surprisingly) in Pollock.

The elimination of identical rows shrinks the partition table, so the time required for each new iteration becomes significantly lower. If $M$ is the size of the image in pixels, the  time complexity of the box merging algorithm is $O(M)$ because it scans the pixels of the colour image only once. In our dual core 32 bits system with Matlab, the time the box merging method needs is 5 seconds per mega-pixel in the worst case.

Apart from the memory required for the colour image itself, the algorithm also needs some memory space for the partition table: 2 bytes for each spatial coordinate and 1 byte for each colour, 7 bytes per pixel in total.

The method presented here is a variation of the box counting method so any evaluation of box counting, except that of time complexity and memory requirements, applies here as well, e.g. \cite{Foroutan-pour1999}, \cite{Theiler1990} and \cite{Peyrin1994}.

\section{Conclusions} \label{sec:Conclusions}
The box merging algorithm presented here, based on box counting, is fast and very easily implemented from one to any number of dimensions. With the exception of time complexity, the presented algorithm has all the characteristics of the box counting method since it produces the same $\log n - \log s$ table. We have tested colour images and found that their measured fractal dimension varies between 2 and 5, validating our algorithm.


\end{document}